# A Review:
# Expert System for Diagnosis of Myocardial Infarction

S. J. Gath
*Lecturer, BCA Department,*
*DKASC College, Ichalkaranji, Kolhapur,*
*Maharashtra*

Dr. R. V. Kulkarni
*Professor & HOD,*
*SIBER, Kolhapur,*
*Maharashtra*

**Abstract:** A computer Program Capable of performing at a human-expert level in a narrow problem domain area is called an expert system. Management of uncertainty is an intrinsically important issue in the design of expert systems because much of the information in the knowledge base of a typical expert system is imprecise, incomplete or not totally reliable.
**In this paper, the author present s the review of past work that has been carried out by various researchers based on development of expert systems for the diagnosis of cardiac disease**

## INTRODUCTION

Expert System is computer software application which has the ability to replicate the thinking and reasoning capacity of human-being based on some facts and rules presented to it. Today expert systems have been used in all most all the fields to assist the users in decision making. They act like experts and provide the expertise decision and assistance for all the fields where human expatriation and complex decision making is required, like medical diagnosis, expert decision making, policy making, estimating strategies, analysis etc. The use of expert system in medical diagnosis started since early 70's when MYCIN, an expert system for identifying bacteria causing diseases was developed at Stanford University. Onwards then, various expert systems like Internist – I, CADUCEUS etc. Have been developed. The goal of such an expert system is to aid the medical experts in making diagnosis of certain diseases or help the ordinary people to diagnose the disease themselves.

Myocardial infarction (MI) or acute myocardial infarction (AMI), commonly known as a heart attack or IM, occurs when one or more coronary arteries become suddenly blocked, resulting in heart muscle death. And results from the interruption of blood supply to a part of the heart, causing heart cells to die. This is most commonly due to occlusion (blockage) of a coronary artery following the rupture of a vulnerable atherosclerotic plaque, which is an unstable collection of lipids (cholesterol and fatty acids) and white blood cells (especially macrophages) in the wall of an artery. The resulting ischemia (restriction in blood supply) and ensuing oxygen shortage, if left untreated for a sufficient period of time, can cause damage or death (infarction) of heart muscle tissue (myocardium).

Expert System for Myocardial infarction is a knowledge-based system capable of making multiple diagnoses in myocardial infarction. Expert System for Myocardial infarction is designed for a personal computer, which can provide assistance in the diagnosis of Myocardial Infarction and can demonstrate the reliability of a microcomputer-based expert system for medical diagnosis in this type of craniological disease.

Today there is need of developing The Expert System for diagnosis of Myocardial infarction because this Heart disease is the leading cause of death today, and causing about million deaths per year in the Society. It is the most common heart diseases and is familiar as "HEART ATTACK", The Cardiology has estimated that there is shortage of specialists, and the worrying fact is that the most of electrocardiograms are analyzed by general practitioners or lab experts. Implementation of The Expert System for Myocardial infarction will aim to have a direct impact on unnecessary visits to the cardiologist by effectively and efficiently decentralizing decision-making, will geared up effective interaction between patients, general practitioners and cardiologists and also this will be useful to train junior doctors and will aim to achieve a satisfactory solution to Physicians and General Practitioner's.

## REVIEW OF PREVIOUS RESEARCH WORK:

**E.P.Ephzibah, Dr. V. Sundarapandian**
**(An International Journal (CSEIJ), Vol.2, No.1, February 2012)**
This paper is based on the heart disease diagnosis in patients with the help of evolutionary techniques like genetic algorithms, Fuzzy rule based expert system and Neural networks. This expert system will help the doctors to arrive at a conclusion about the presence or absence of heart disease in patients. This is an enhanced system that accurately classifies the presence of the heart disease. It is proved that the generated neuro-fuzzy rule based classification system is capable of diagnosing the heart disease in an effective and efficient method than any other classifiers.

**Tilotma Sharma, , Sumit Jain:**
 (ISSN 2250-2459, Volume 2, Issue 1, January 2012)
In this paper, biomedical, agricultural, and automobile areas come under diagnosis and troubleshooting using ES that performs all the necessary data manipulation as a human expert would do. Others areas uses different techniques like fuzzy logic, rule-based reasoning, CLIPS etc. which actually comes under decision and planning area of ES. No matter which area of business one is engaged in, ES can fulfill the need of higher productivity and reliability of decisions. Everyone can find an application potential in the field of ESs. Expert Systems can actually help to create opportunities for new job areas. In order to be considered useful and acceptable ESs must be able to explain their knowledge of the domain and the reasoning processes they employ to produce results and recommendations. Expert systems are an example of symbolic paradigm, the other major paradigm is the connectionist paradigm that has led to the development of Neural Network Technology.

**Ahmad A. Al-Hajji:**
 (ICCIT 2012)
Abstract- This paper present an account of Rule-Based Expert System (RBES) for Neurological Disorders, i.e., Alzheimer, Parkinson, Huntington's disease, Cerebral Palsy, Meningitis, Epilepsy, Multiple Sclerosis, Stroke, Cluster headache, Migraine, Meningitis for children. More than10 types of neurological diseases can be diagnosed and treated by this system. It is rule-based expert system, assisting Neurologists, medical students doing specialization in neurology, researchers as well as neurological patients having computer know-how. It helps the patients to get the required advice about the different disorders attack to them due to their nervous system disorders. The expert rules were developed on the symptoms of each type of neurological disease, and they were presented using decision tree and inferred using backward-chaining method. The knowledge base consists of information, collected from books and doctors (domain experts) about neurology and its disorders. Declarative and conceptual system models were developed with Prolog and MS Visual Studio.Net technologies

**Mohammed Aslam.C, SatyaNarayana.D, Padma Priya.K :**

(International Journal of Computer Applications in Engineering Sciences [VOL I, ISSUE IV, DECEMBER 2011])
In this paper, a novel fuzzy inference system and mathematical morphologic algorithms is proposed to segmentation of blood vessels. The results show that the algorithm is more efficient for segmentation of
angiogram images and noise cancelling than other methods such as canny method

**Djam, and Y.H. Kimbi:**
(volume 12. Number 1. May 2011 (Spring))
This paper focused on the use of information and communication technology (ICT) to design a web-based fuzzy expert system for the management of hypertension using the fuzzy logic approach. In this paper, systolic blood pressure, diastolic blood pressure, age, and body mass index (BMI) were taken as input parameters to the fuzzy expert system and "hypertension risk" was the output parameter. The resultant hypertension risk was based on fuzzy rules that were developed for the expert system. The input triangular membership functions are Low, Normal, High, and Very High for blood pressure. The output triangular membership functions are mild, moderate and severe. The defuzzification method used in this
paper is the Root Sum Square. The expert system was designed based on clinical observations, medical diagnosis, and the experts' knowledge. The expert system provides a web-based interface that was designed using PHP as a scripting language with MySQL relational database on Apache Sever under Windows operating system platform, and using Java Script and HTML to archive the system design. Unified Modeling Language was used to describe the logical design of the system. We selected 50 patients with hypertension and computed the results that were in the range of predefined limit by the domain experts.

**Mir Anamul Hasan, Khaja Md. Sher-E-Alam and Ahsan Raja Chowdhury**
(JOURNAL OF COMPUTING, VOLUME 2, ISSUE 6, JUNE 2010, ISSN2151-9617 )
Human disease diagnosis is a complicated process and requires high level of expertise. Any attempt of developing a web-based expert system dealing with human disease diagnosis has to overcome various difficulties. This paper describes a project work aiming to develop a web-based fuzzy expert system for diagnosing human diseases. Now a days fuzzy systems are being used successfully in an increasing number of application areas; they use linguistic rules to describe systems. This research project focuses on the research and development of a web-based clinical tool designed to improve the quality of the exchange of health information between health care professionals and patients. Practitioners can also use this web-based tool to corroborate diagnosis. The proposed system is experimented on various scenarios in order to evaluate it's performance.

**C. Angeli**
(Advanced Knowledge Based Systems: Model, Applications & Research (Eds. Sajja & Akerkar), Vol. 1, pp 50 – 73, 2010 )
In this paper, the evolution of expert systems paradigm for the solutions of the diagnostic problem have been presented as well as the evolution of knowledge acquisition, representation and user interface methods for the expert systems development process. Experiential knowledge, scientific knowledge and a combination of the two sources of knowledge has been used to perform the diagnostic task

**Shaikh Abdul Hannan, V. D. Bhagile, R. R. Manza,, R. J. Ramteke :**
**(Vol. 02, No. 06, 2010, 2150-2159)**
In this paper, around 300 patient's information is collected from Sahara Hospital, under supervision of Dr. Abdul Jabbar, (MD Medicine) Sahara Hospital, Roshan Gate, Aurangabad. The collected information is coded, normalized and entered into 13 different excel sub-sheets. All the patients data is trained by using SVM and FFBP. Around 50 samples were tested with these two techniques. If the more data set is used for the training the NN model gives more robust results. The analysis model by using SVM and FFBPof ANN gives less appropriate result for medical prescription for heart disease patient. However, there are several techniques that can improve the speed and performance of the back propagation algorithm, weight initialization, use of momentum and adaptive learning rate. It is found that the result of testing data by using SVM and FFBP is not satisfactory as per the result verified by the doctor.

**Ali.Adeli, Mehdi.Neshat:**
**(IMECS , March 17-2010 Hong Kong)**
The designed system based on the V.A. Medical Center, Long Beach and Cleveland Clinic Foundation data base. The system has 13 input fields and one output field. Input fields are chest pain type, blood pressure, cholesterol, resting blood sugar, maximum heart rate, resting electrocardiography (ECG), exercise, old peak (ST depression induced by exercise relative to rest), thallium scan, sex and age. The output field refers to the presence of heart disease in the patient. It is integer valued from 0 (no presence) to 4 (distinguish presence (values 1, 2, 3, 4)). This system uses Mamdani inference method. The results obtained from designed system are compared with the data in upon database and observed results of designed system are correct in 94%. The system designed in Matlab software. The system can be viewed as an alternative

**Shaikh Abdul Hannan, V. D. Bhagile, Lecturer, R. R. Manza,R. J. Ramteke :**
**((IJCSE) International Journal on Computer Science and Engineering Vol. 02, No. 06, 2010, 2150-2159)**
In this paper It is found that the result of testing data by using SVM is not satisfactory but the medicines prescribed by the RBF are satisfactory as per the result verified by the doctor. The diagnosis performances of this study shows the advantage of this system. : it is rapid, easy to operate, non-invasive and not expensive. The working prototype model in the field of heart diagnosis can use the system. It also helps for training beginner's doctors and medical students who work in the field of heart diagnosis

**V.Mahesh, A.Kandaswamy, and R.Venkatesan**
**(nternational Journal of Engineering and Technology Vol. 1, No.3, August, 2009 ISSN: 1793-8236):**
This paper presents the development of a rule-based expert system that emulates the ECG interpretation skills of an expert cardiologist. The knowledge of an expert is confined to him and is not freely available for decision-making. An expert system is developed to overcome this problem. In this rule-based expert system, patient's heart rate and the wave characteristics of the ECG are considered. With these 'facts', rules are framed and a rule base is developed in consultation with experts. An inference engine in the expert system uses these inputs and the rule base to identify any abnormality in the patient's heart. A unique feature of the proposed system is storing the post diagnostic information in a database. That is, information about a patient diagnosed by this system is stored and can be retrieved later date-wise. The user can also input his comments or suggestions for improvement or correction, which can later be incorporated in the system after validation by experts. Only authorized access to the database is permitted. This expert system can support physicians in their diagnosis and decision-making.

**V.Mahesh, A.Kandaswamy, and R.Venkatesan:**
**(International Journal of Recent Trends in Engineering, Vol 2, No. 3, November 2009)**
his paper presents the development of a web based telecardiology framework for the diagnosis of cardiac patients in rural areas. Rural health centers, a centralized server and expert cardiologists constitute the framework. Personal details, clinical information and ECG acquired from a patient at the rural center are sent to a centralized server. The server processes the ECG file to extract the lead measurements and the waveform data. The expert views the patient's clinical information, lead measurements and ECG waveforms and submits his diagnosis for treatment. The server stores the diagnosed results and forwards the same to the respective rural center. The physician at the rural center can view the diagnosed results and recommendations from the expert and effect treatment.

**Fahad Shahbaz Khan , Saad Razzaq, Kashif Irfan, Fahad Maqbool, Ahmad Farid, Inam Illahi, Tauqeer ul amin :**
**(WCE 2008 , July 2-4-2008, London , U.K.)**
This paper presents a web-based expert system for wheat crop in Pakistan. Wheat is one of the major grain crops in Pakistan. It is cultivated in vast areas of Punjab followed by Sindh and ranked first as a cereal crop in the country. this rule-based expert system covers two main classes of problems namely diseases and pests, normally encountered in wheat crop. The expert system is intended to help the farmers, researchers and students and provides an efficient and goal-oriented approach for solving common problems of wheat. The system gives results that are correct and consistent.

**Markos G. Tsipouras, Dimitrios I. Fotiadis**
**(IEEE TRANSACTIONS ON BIOMEDICAL ENGINEERING, VOL. 54, NO. 11, NOVEMBER 2007)**
In the proposed methodology, the initial crisp set of rules are determined by experts. Starting from a crisp set of rules and then transforming it into a fuzzy model allows this methodology to be applied in cases where the initial set of rules is strictly crisp. Based on this feature, an alternative is to extract rules from the data, using a data mining technique, instead of a knowledge-based origin of the initial set of rules. In this case, the methodology would be fully automated, data driven, but the knowledge introduced from the experts through the initial set of rules, would be lost. Furthermore, hybrid approaches, combining to expert's knowledge and data-mined rules are also applicable. In this context, also the ability to automatically predetermine some of the fuzzy model's basic aspects, such as the fuzzy membership function and the $T_{norm}$ and $S_{norm}$ definitions based on the natural characteristics of the problem, is a significant field of research. Another important feature is that the gradient is available for some of the fuzzy models; this feature can lead to the use of more efficient optimization methods, which take advantage of the first derivative information.

**Amani Al-Ajlan : April – 2007**
**(KING SAUD UNIVERSITY ,COLLEGE OF COMPUTER & INFORMATION SCIENCES ,DEPARTMENT OF COMPUTER SCIENCE ,Second Semester 1427/1428 AH)**
In this report the author review two medical expert systems HDP and PUFF. They assist physicians in daily work of the diagnosis of patients and they spread the expert knowledge of a few highly skilled advanced doctors leading experts in their fields to a much broader medical staff. The author thinks medical expert system provide a challenging domain in artificial intelligent and need more study and research especially of it accuracy effect human life.

**Miss Hong Liang**
**(Proceedings of the 2005 IEEE Engineering in Medicine and Biology 27th Annual Conference Shanghai, China, September 1-4, 2005)**
This paper proposed ECG Feature Elements Identification method for Cardiologist Expert Diagnosis. This method includes ECG detection, Noise Purification, ECG PQRST Key Features identification even they are confused with each other or incomplete, the key parameters (the amplitudes, duration, areas, shapes, intervals and slopes) estimation, Global ECG Data transfer by email, Logic Category Classification, Measurements Summary, and Computer-generated ECG Interpretation. This method is explicit, highly efficient, highly accurate, and reliable. It can run in real time and can offer a clear and explicit explanation for the Cardiologist's diagnosis. Future work will focus on Global ECG index system, ECG Logic Category Classification and logic linking memory, and Cardiologist Expert's Decision-Rule (and Knowledge) Abstract For Exact Expert Diagnosis.

**Abdel-Badeeh M. Salem, Mohamed Roushdy and Rania A. HodHod :**
**(AIML Journal, Volume (5), Issue (1), March, 2005)**
In this paper, a case-based medical expert system prototype that supports diagnosis of four heart diseases was developed. Knowledge structure was represented via a formalism of cases. The system used two different techniques for the retrieval process; induction and nearest-neighbor techniques. The later technique was adapted to K-nearest neighbor to overcome the problem of obtaining more than one diagnosis for an input case. The results indicate that the nearest neighbor is better than the induction strategy, where the retrieval accuracy were100% and 53.8% respectively The nearest neighbor strategy was verified by conducting several tests in order to measure retrieval accuracy, retrieval consistency, case duplication and the overall system performance. Cardiologists evaluate the system performance by testing it practically for 13 new cases where the system succeeded in estimating the correct diagnosis. For future work, more cases will be added to the case memory and it will be clinically tested.

**Marcin Grabowski, Krzysztof J. Filipiak , Robert Rudowski, Grzegorz Opolski :**
**(Pol J Pathol 2003, 54, 3, 205-208)**
In this paper The verification based on archival data has confirmed the reliability of the proposed expert system. The differences between physicians' decisions and ES advice may be attributed to changes of guidelines. The expert system has to be continuously updated according to new guidelines and recommendations. differences classes of recommendations must be considered.

**Costas Papaloukas, Dimitrios I. Fotiadis, Aristidis Likas, Christos S. Stroumbis and Lampros K. Michalis**
**(Journal of Electrocardiology Vol. 35 No. 1 2002)**
This new rule-based expert system rapidly and reliably detects ECG changes suggestive of ischemia in long duration ECGs. Its performance is not affected by the presence of noise and clarifies the type of each detected episode. Whether the advantages of this new system will increase the diagnostic and prognostic accuracy requires further evaluation.

**William J. Long: 1992**
This paper reports on a formative evaluation of the diagnostic capabilities of the Heart Failure Program, which uses a probability network and a heuristic hypothesis generator. Using 242 cardiac cases collected from discharge summaries at a tertiary care hospital, we compared the diagnoses of the program to diagnoses collected from cardiologists using the same information as was available to the program. With some adjustments to the knowledge base, the Heart Failure Program produces appropriate diagnoses about 90% of the time on this training set. The main reasons for the inappropriate diagnoses of the remaining 10% include inadequate reasoning with temporal relations between cause and effect, severity relations, and independence of acute and chronic diseases.

**Shu-hsien Liao :**
This paper is based on a literature review on Knowledge Management technologies and applications from 1995 to 2002 using a keyword index search. We conclude that KM technologies tend to develop towards expert orientation, and KM applications development is a problem-oriented domain. Different social studies methodologies, such as statistical method, are suggested to implement in KM as another kind of technology. Integration of qualitative and quantitative methods, and integration of KM technologies studies may broaden the horizon on this subject. Finally, the ability to continually change and obtain new understanding is the power of KM technologies and will be the application of future works.

**AHMED A. RADWAN, HAZEM M. EL-BAKRY and HAGER M. EL HADAD**
**(SBN: 978-1-61804-034-3)**
In the presented study they have trained the network with large amount of data and they see that as they train the network, the network can give correct decision about the correct examination to the entered symptoms. As they know that the study is about pediatric respiratory disease. They also see that the network learning way simulation of human learn manner. Human learn depends on the amount of cases that they face. Artificial neural networks also learn after it faces 699 cases with different symptoms. The computer program was performed under MATLAB software using the neural network toolbox. In the training, the number of neurons on the hidden layer is 20. A dataset including 699 data samples obtained from experimental studies were used for ANNs. From these, 699 data patterns were used for training the network, and the remaining 20% patterns were randomly selected and used as the test dataset. The results are shown in Figure. After training the network they test it with some different cases with out the target output vector and the network gives us the correct diagnosis. So the artificial neural network is very useful data mining technique for recognition of pediatric respiratory disease.

**MARCIN GRABOWSKI, KRZYSZTOF J. FILIPIAK,ROBERT RUDOWSKI1 AND GRZEGORZ OPOLSKI:**
The aim of the project was to create a computer program expert system, which will support a doctor when a management for patients with acute coronary syndrome needs to be chosen. The expert system consists of four modules: knowledge base, previous cases database, inference engine and explanation module. Knowledge base was created with support of clinical experts, based on current management standards, guidelines and results of clinical trials according to evidence-based medicine rules. Data from new patient are added to the case database. Inference engine integrates two types of reasoning rule-based and case-based reasoning. Computer expert system gives unambiguous and objective answer. Recommendation given by an expert system can be reliable. At present the system is tested in clinical practice. Strategies recommended by the system are compared to the management applied in patients treated in Cardiology Clinic.

**G.N.R. Prasad, Dr. A. Vinaya Babu:**
This paper has discussed the need of expert systems in agriculture and availability of various expert systems in various countries. The need of expert systems for technical information transfer in agriculture can be identified by recognizing the problems. The advantages that
an expert system can offer better    than  traditional methods. It is proven that expert    systems in agriculture helps a   lot in increasing the  crop   production. But most of the expert systems are in    English language.   By developing an expert system   in agriculture   in a mother tongue of a farmer, helps  him/her   to know the facts and truths in increasing the production.

**Yasser Abdelhamid, Hesham Hassan, Ahmed Rafea :**

**Paul Dorsey, Dulcian :**
The methodology presented in this paper ties three main aspects: theoretical basis, practical implementation, and workflow organization. The theoretical basis are derived from CommonKADS, the second generation expert system approaches state of the art. The practical experience was gained through the development of several expert systems in the domain of crop production management, this directly affected both the methodology application, and the workflow organizational structure of the lab

**M Sasikumar,S Ramani,S Muthu Raman KSR Anjaneyulu,R Chandrasekar :**
There are some basic ideas in expert systems which one should understand before implementing expert systems. However, these ideas should only be the stepping stones to the practical aspects of expert systems technology. Keeping this in mind, we have tried to provide a judicious mixture of principles and practical aspects in this book. The book is structured around a part-time course on expert systems which is conducted at the National Centre for Software Technology (NCST). The emphasis in that course, and in this book, is on rule based expert systems. However, aspects related to other types of expert systems have also been covered

**Vanisree K , yothi Singaraju :**
his paper presents a Decision Support System for Congenital Heart Disease diagnosis which can speed up diagnosis process, improves the accuracy of the diagnosis and reduces the cost without loss of diagnostic performance. The proposed system also decreases the number of diagnostic tests and predicts the disease at an early stage of the disease. Therefore the present system gives a better performance when comparing with the manual diagnosis

**Romeo Mark A. Mateo, Bobby D. Gerardo and Jaewan Lee**
This paper presents a healthcare expert system (HES) based group cooperation model to implement QoS and efficiency of the system by using object replicas and effective coordination of the system using the proposed agent manager. The group cooperation model was discussed for the proposed healthcare expert system. A group of objects or service is managed by the proposed agent manager where it chooses the necessary objects from the replica server. The global load monitor and analyzer manage the flow of the adaptive scheme to implement the efficient distribution of loads. The replica manager handles the replication of the objects. The adaptive load distribution is based on two algorithms which are the round robin and fuzzy least load algorithm. This research implements the group cooperation model for the HES implementing the efficient service by using components and objects interactions and provides QoS by using replication and adaptive load distribution schemes

**Rajdeep Borgohain and Sugata Sanyal**
In this paper we have discussed the design and implementation of a rule based Expert System for Cerebral Palsy Diagnosis. The expert system helps to diagnose Cerebral Palsy and classify it as mild, moderate or severe. In the implementation, we have taken the most classical symptoms of Cerebral Palsy and given a weightage to each of the symptom and according to the feedback given by the user. The expert system can go a great deal in supporting the decision making process of medical professionals and also help parents having children with Cerebral Palsy to assess their children and to take appropriate measures to manage the disease.

**Eric J. Horvitz :**
Pathfinder is an expert system that assists surgical pathologists with the diagnosis of lymph-node diseases. The program is one of a growing number of normative expert systems that use probability and decision theory to acquire, represent, manipulate, and explain uncertain medical knowledge. In this article, we describe Pathfinder and the research in uncertain-reasoning paradigms that was stimulated by the development of the program. We discuss limitations with early decision-theoretic methods for reasoning under uncertainty and the initial attempts to use non-decision-theoretic methods. Then, we describe experimental and theoretical results that directed us to return to reasoning methods based in probability and decision theory.

**Rahime Ceylan , Yüksel Özbay , Bekir Karlik :**
In this paper, the new telemedicine system as automated diagnostic system assisted to practitioner doctor was presented. Furthermore, the T2FCNN was proposed and developed to classify electrocardiography signals. In proposed structure, the conventional type-1 memberships for each pattern were extended to type-2. These type-2 memberships were incorporated in the cluster updating process. The cluster centers obtained by T2FCM are classified by neural network. The realized T2FCNN structure was compared with the studies in literature. It can be seen in Table 4 that the most high recognition rate was obtained using T2FCNN for five ECG class types.

## CONCLUSION

In this paper the author has tried to present a detailed review of previous work carried out by various researchers in the field of development of expert system for diagnosis of various diseases related to cardiology and some other expert systems related to diagnosis of human diseases.

And the author has came out with conclusion that expert system can be able to capture expert knowledge before it is lost, the Expert system can help in decision making, expert system can also assist junior doctors and Practitioner's in complicated diagnosis, the expert system can impart the real time knowledge for automation in process control, the expert system can be able to diagnose a disease bases on series of questions

With this conclusion the author has decided to develop an expert system for diagnosis of Myocardial Infarction as the author has found that there has been no any such work carried previously.